# PWC-Net: CNNs for Optical Flow Using Pyramid, Warping, and Cost Volume

Deqing Sun, Xiaodong Yang, Ming-Yu Liu, and Jan Kautz
NVIDIA

## Abstract

*We present a compact but effective CNN model for optical flow, called PWC-Net. PWC-Net has been designed according to simple and well-established principles: pyramidal processing, warping, and the use of a cost volume. Cast in a learnable feature pyramid, PWC-Net uses the current optical flow estimate to warp the CNN features of the second image. It then uses the warped features and features of the first image to construct a cost volume, which is processed by a CNN to estimate the optical flow. PWC-Net is 17 times smaller in size and easier to train than the recent FlowNet2 model. Moreover, it outperforms all published optical flow methods on the MPI Sintel final pass and KITTI 2015 benchmarks, running at about 35 fps on Sintel resolution ($1024 \times 436$) images. Our models are available on* https://github.com/NVlabs/PWC-Net.

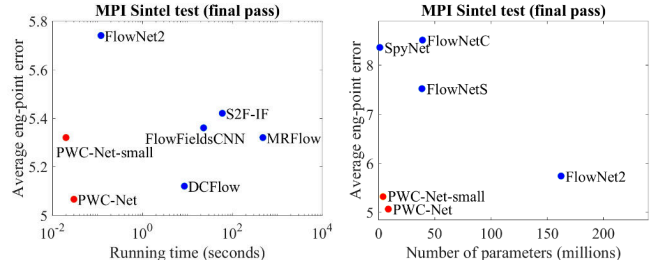

Figure 1. Left: PWC-Net outperforms all published methods on the MPI Sintel final pass benchmark in both accuracy and running time. Right: among existing end-to-end CNN models for flow, PWC-Net reaches the best balance between accuracy and size.

## 1. Introduction

Optical flow estimation is a core computer vision problem and has many applications, *e.g.*, action recognition [44], autonomous driving [26], and video editing [8]. Decades of research efforts have led to impressive performances on challenging benchmarks [4, 12, 18]. Most top-performing methods adopt the energy minimization approach introduced by Horn and Schunck [19]. However, optimizing a complex energy function is usually computationally expensive for real-time applications.

One promising approach is to adopt the fast, scalable, and end-to-end trainable convolutional neural network (CNN) framework [31], which has largely advanced the field of computer vision in recent years. Inspired by the successes of deep learning in high-level vision tasks, Dosovitskiy *et al.* [15] propose two CNN models for optical flow, *i.e.*, FlowNetS and FlowNetC, and introduce a paradigm shift. Their work shows the feasibility of directly estimating optical flow from raw images using a generic U-Net CNN architecture [40]. Although their performances are below the state of the art, FlowNetS and FlowNetC models are the best among their contemporary real-time methods.

Recently, Ilg *et al.* [24] stack several FlowNetC and FlowNetS networks into a large model, called FlowNet2, which performs on par with state-of-the-art methods but runs much faster (Fig. 1). However, large models are more prone to the over-fitting problem, and as a result, the subnetworks of FlowNet2 have to be trained sequentially. Furthermore, FlowNet2 requires a memory footprint of 640MB and is not well-suited for mobile and embedded devices.

SpyNet [38] addresses the model size issue by combining deep learning with two classical optical flow estimation principles. SpyNet uses a spatial pyramid network and warps the second image toward the first one using the initial flow. The motion between the first and warped images is usually small. Thus SpyNet only needs a small network to estimate the motion from these two images. SpyNet performs on par with FlowNetC but below FlowNetS and FlowNet2 (Fig. 1). The results by FlowNet2 and SpyNet show a clear trade-off between accuracy and model size.

*Is it possible to both increase the accuracy and reduce the size of a CNN model for optical flow?* In principle, the trade-off between model size and accuracy imposes a fundamental limit for general machine learning algorithms. However, we find that combining domain knowledge with deep learning can achieve both goals simultaneously.

SpyNet shows the potential of combining classical principles with CNNs. However, we argue that its performance gap with FlowNetS and FlowNet2 is due to the partial use of the classical principles. First, traditional optical flow methods often pre-process the raw images to extract features that are invariant to shadows or lighting changes [4, 48]. Fur-ther, in the special case of stereo matching, a cost volume is a more discriminative representation of the disparity (1D flow) than raw images or features [20, 42, 59]. While constructing a full cost volume is computationally prohibitive for real-time optical flow estimation [55], this work constructs a "partial" cost volume by limiting the search range at each pyramid level. We can link different pyramid levels using a warping layer to estimate large displacement flow.

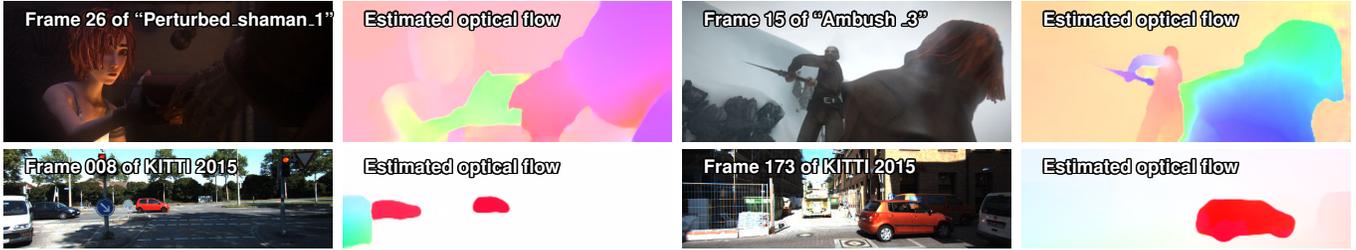

Figure 2. PWC-Net results on Sintel final pass (top) and KITTI 2015 (bottom) test sets. It outperforms all published flow methods to date.

Our network, called PWC-Net, has been designed to make full use of these simple and well-established principles. It makes significant improvements in model size and accuracy over existing CNN models for optical flow (Figs. 1 and 2). At the time of writing, PWC-Net outperforms all published flow methods on the MPI Sintel final pass and KITTI 2015 benchmarks. Furthermore, PWC-Net is about 17 times smaller in size and provides 2 times faster inferencing than FlowNet2. It is also easier to train than SpyNet and FlowNet2 and runs at about 35 frames per second (fps) on Sintel resolution (1024×436) images.

## 2. Previous Work

**Variational approach.** Horn and Schunck [19] pioneer the variational approach to optical flow by coupling the brightness constancy and spatial smoothness assumptions using an energy function. Black and Anandan [7] introduce a robust framework to deal with outliers, *i.e.*, brightness inconstancy and spatial discontinuities. As it is computationally impractical to perform a full search, a coarse-to-fine, warping-based approach is often adopted [11]. Brox *et al.* [9] theoretically justify the warping-based estimation process. Sun *et al.* [45] review the models, optimization, and implementation details for methods derived from Horn and Schunck and propose a non-local term to recover motion details. The coarse-to-fine, variational approach is the most popular framework for optical flow. However, it requires solving complex optimization problems and is computationally expensive for real-time applications.

One conundrum for the coarse-to-fine approach is small and fast moving objects that disappear at coarse levels. To address this issue, Brox and Malik [10] embed feature matching into the variational framework, which is further improved by follow-up methods [50, 56]. In particular, the EpicFlow method [39] can effectively interpolate sparse matches to dense optical flow and is widely used as a post-processing method [1, 3, 14, 21, 55]. Zweig and Wolf [60] use CNNs for sparse-to-dense interpolation and obtain consistent improvement over EpicFlow.

Most top-performing methods use CNNs as a component in their system. For example, DCFlow [55], the best published method on MPI Sintel final pass so far, learns CNN features to construct a full cost volume and uses sophisticated post-processing techniques, including EpicFlow, to estimate the optical flow. The next-best method, FlowFieldsCNN [3], learns CNN features for sparse matching and densifies the matches by EpicFlow. The third-best method, MRFlow [53] uses a CNN to classify a scene into rigid and non-rigid regions and estimates the geometry and camera motion for rigid regions using a plane + parallax formulation. However, none of them are real-time or end-to-end trainable.

**Early work on learning optical flow.** Simoncelli and Adelson [43] study the data matching errors for optical flow. Freeman *et al.* [16] learn parameters of an MRF model for image motion using synthetic blob world examples. Roth and Black [41] study the spatial statistics of optical flow using sequences generated from depth maps. Sun *et al.* [46] learn a full model for optical flow, but the learning has been limited to a few training sequences [4]. Li and Huttenlocker [32] use stochastic optimization to tune the parameters for the Black and Anandan method [7], but the number of parameters learned is limited. Wulff and Black [52] learn PCA motion basis of optical flow estimated by GPU-Flow [51] on real movies. Their method is fast but produces over-smoothed flow.

**Recent work on learning optical flow.** Inspired by the success of CNNs on high-level vision tasks [29], Dosovitskiy *et al.* [15] construct two CNN networks, FlowNetS and FlowNetC, for estimating optical flow based on the U-Net denoising autoencoder [40]. The networks are pre-trained on a large synthetic FlyingChairs dataset but can surprisingly capture the motion of fast moving objects on the Sintel dataset. The raw output of the network, however, contains large errors in smooth background regions and requires variational refinement [10]. Mayer *et al.* [35] apply the FlowNet architecture to disparity and scene flow esti-mation. Ilg *et al.* [24] stack several basic FlowNet models into a large one, *i.e.*, FlowNet2, which performs on par with state of the art on the Sintel benchmark. Ranjan and Black [38] develop a compact spatial pyramid network, called SpyNet. SpyNet achieves similar performance as the FlowNetC model on the Sintel benchmark, which is good but not state-of-the-art.

Another interesting line of research takes the unsupervised learning approach. Memisevic and Hinton [36] propose the gated restricted Boltzmann machine to learn image transformations in an unsupervised way. Long *et al.* [34] learn CNN models for optical flow by interpolating frames. Yu *et al.* [58] train models to minimize a loss term that combines a data constancy term with a spatial smoothness term. While inferior to supervised approaches on datasets with labeled training data, existing unsupervised methods can be used to (pre-)train CNN models on unlabeled data [30].

**Cost volume.** A cost volume stores the data matching costs for associating a pixel with its corresponding pixels at the next frame [20]. Its computation and processing are standard components for stereo matching, a special case of optical flow. Recent methods [14, 15, 55] investigate cost volume processing for optical flow. All build the full cost volume at a single scale, which is both computationally expensive and memory intensive. By contrast, our work shows that constructing a partial cost volume at multiple pyramid levels leads to both effective and efficient models.

**Datasets.** Unlike many other vision tasks, it is extremely difficult to obtain ground truth optical flow on real-world sequences. Early work on optical flow mainly relies on synthetic datasets [5], *e.g.*, the famous "Yosemite". Methods may over-fit to the synthetic data and do not perform well on real data [33]. Baker *et al.* [4] capture real sequences under both ambient and UV lights in a controlled lab environment to obtain ground truth, but the approach does not work for outdoor scenes. Liu *et al.* [33] use human annotations to obtain ground truth motion for natural video sequences, but the labeling process is time-consuming.

KITTI and Sintel are currently the most challenging and widely-used benchmarks for optical flow. The KITTI benchmark is targeted for autonomous driving applications and its semi-dense ground truth is collected using LIDAR [18]. The 2012 set only consists of static scenes. The 2015 set is extended to dynamic scenes via human annotations and more challenging to existing methods because of the large motion, severe illumination changes, and occlusions [37]. The Sintel benchmark [12] is created using the open source graphics movie "Sintel" with two passes, clean and final. The final pass contains strong atmospheric effects, motion blur, and camera noise, which cause severe problems to existing methods. All published, top-performing methods [3, 53, 55] rely heavily on traditional techniques. By embedding the classical principles into the network architecture, we show that a fully end-to-end method can outperform all published methods on both the KITTI 2015 and Sintel final pass benchmarks.

**CNN models for dense prediction tasks in vision.** The denoising autoencoder [47] has been commonly used for dense prediction tasks in computer vision, especially with skip connections [40] between the encoder and decoder. Recent work [13, 57] shows that dilated convolution layers can better exploit contextual information and refine details for semantic segmentation. Here we use dilated convolutions to integrate contextual information for optical flow and obtain moderate performance improvement. The DenseNet architecture [22, 27] directly connects each layer to every other layer in a feedforward fashion and has been shown to be more accurate and easier to train than traditional CNN layers in image classification tasks. We test this idea for dense optical flow prediction.

## 3. Approach

Figure 3 summarizes the key components of PWC-Net and compares it side by side with the traditional coarse-to-fine approach [7, 9, 19, 45]. First, as raw images are variant to shadows and lighting changes [9, 45], we replace the fixed image pyramid with learnable feature pyramids. Second, we take the warping operation from the traditional approach as a layer in our network to estimate large motion. Third, as the cost volume is a more discriminative representation of the optical flow than raw images, our network has a layer to construct the cost volume, which is then processed by CNN layers to estimate the flow. The warping and cost volume layers have no learnable parameters and reduce the model size. Finally, a common practice by the traditional methods is to post-process the optical flow using contextual information, such as median filtering [49] and bilateral filtering [54]. Thus PWC-Net uses a context network to exploit contextual information to refine the optical flow. Compared with energy minimization, the warping, cost volume, and CNN layers are computationally light.

Next, we will explain the main ideas for each component, including pyramid feature extractor, optical flow estimator, and context networks. Please refer to the supplementary material for details of the networks.

**Feature pyramid extractor.** Given two input images $\mathbf{I}_1$ and $\mathbf{I}_2$, we generate $L$-level pyramids of feature representations, with the bottom (zeroth) level being the input images, *i.e.*, $\mathbf{c}_t^0 = \mathbf{I}_t$. To generate feature representation at the $l$th layer, $\mathbf{c}_t^l$, we use layers of convolutional filters to downsample the features at the $l{-}1$th pyramid level, $\mathbf{c}_t^{l-1}$, by a factor of 2. From the first to the sixth levels, the number of feature channels are respectively 16, 32, 64, 96, 128, and 196.

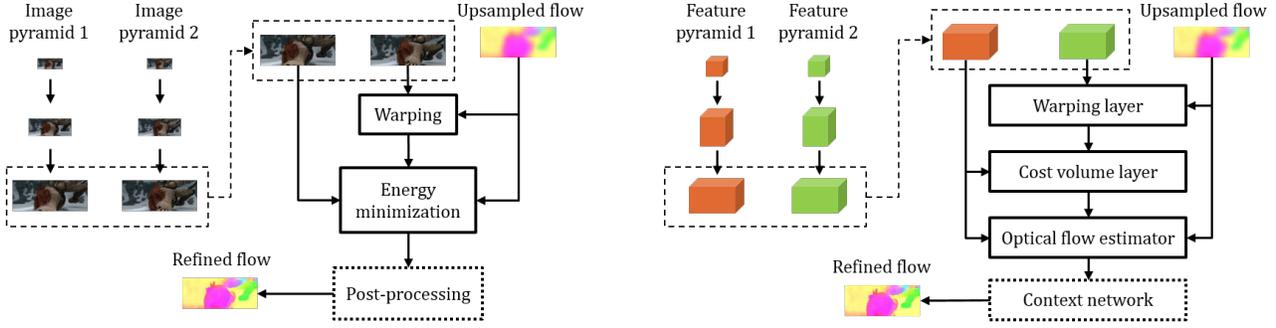

Figure 3. **Traditional coarse-to-fine approach vs. PWC-Net.** Left: Image pyramid and refinement at one pyramid level by the energy minimization approach [7, 9, 19, 45]. Right: Feature pyramid and refinement at one pyramid level by PWC-Net. PWC-Net warps features of the second image using the upsampled flow, computes a cost volume, and process the cost volume using CNNs. Both post-processing and context network are optional in each system. The arrows indicate the direction of flow estimation and pyramids are constructed in the opposite direction. Please refer to the text for details about the network.

**Warping layer.** At the $l$th level, we warp features of the second image toward the first image using the $\times 2$ upsampled flow from the $l+1$th level:

$$\mathbf{c}_w^l(\mathbf{x}) = \mathbf{c}_2^l\big(\mathbf{x} + \text{up}_2(\mathbf{w}^{l+1})(\mathbf{x})\big), \quad (1)$$

where $\mathbf{x}$ is the pixel index and the upsampled flow $\text{up}_2(\mathbf{w}^{l+1})$ is set to be zero at the top level. We use bilinear interpolation to implement the warping operation and compute the gradients to the input CNN features and flow for backpropagation according to [24, 25]. For non-translational motion, warping can compensate for some geometric distortions and put image patches at the right scale.

**Cost volume layer.** Next, we use the features to construct a cost volume that stores the matching costs for associating a pixel with its corresponding pixels at the next frame [20]. We define the matching cost as the correlation [15, 55] between features of the first image and warped features of the second image:

$$\mathbf{cv}^l(\mathbf{x}_1, \mathbf{x}_2) = \frac{1}{N}\big(\mathbf{c}_1^l(\mathbf{x}_1)\big)^\mathsf{T} \mathbf{c}_w^l(\mathbf{x}_2), \quad (2)$$

where $\mathsf{T}$ is the transpose operator and $N$ is the length of the column vector $\mathbf{c}_1^l(\mathbf{x}_1)$. For an $L$-level pyramid setting, we only need to compute a partial cost volume with a limited range of $d$ pixels, i.e., $|\mathbf{x}_1-\mathbf{x}_2|_\infty \leq d$. A one-pixel motion at the top level corresponds to $2^{L-1}$ pixels at the full resolution images. Thus we can set $d$ to be small. The dimension of the 3D cost volume is $d^2 \times H^l \times W^l$, where $H^l$ and $W^l$ denote the height and width of the $l$th pyramid level, respectively.

**Optical flow estimator.** It is a multi-layer CNN. Its input are the cost volume, features of the first image, and upsampled optical flow and its output is the flow $\mathbf{w}^l$ at the $l$th level. The numbers of feature channels at each convolutional layers are respectively 128, 128, 96, 64, and 32, which are kept fixed at all pyramid levels. The estimators at different levels have their own parameters instead of sharing the same parameters. This estimation process is repeated until the desired level, $l_0$.

The estimator architecture can be enhanced with DenseNet connections [22]. The inputs to every convolutional layer are the output of and the input to its previous layer. DenseNet has more direct connections than traditional layers and leads to significant improvement in image classification. We test this idea for dense flow prediction.

**Context network.** Traditional flow methods often use contextual information to post-process the flow. Thus we employ a sub-network, called the context network, to effectively enlarge the receptive field size of each output unit at the desired pyramid level. It takes the estimated flow and features of the second last layer from the optical flow estimator and outputs a refined flow.

The context network is a feed-forward CNN and its design is based on dilated convolutions [57]. It consists of 7 convolutional layers. The spatial kernel for each convolutional layer is $3\times 3$. These layers have different dilation constants. A convolutional layer with a dilation constant $k$ means that an input unit to a filter in the layer are $k$-unit apart from the other input units to the filter in the layer, both in vertical and horizontal directions. Convolutional layers with large dilation constants enlarge the receptive field of each output unit without incurring a large computational burden. From bottom to top, the dilation constants are $1, 2, 4, 8, 16, 1$, and $1$.

**Training loss.** Let $\Theta$ be the set of all the learnable parameters in our final network, which includes the feature pyramid extractor and the optical flow estimators at different pyramid levels (the warping and cost volume layers have no learnable parameters). Let $\mathbf{w}_\Theta^l$ denote the flow field at the $l$th pyramid level predicted by the network, and $\mathbf{w}_{\text{GT}}^l$ the corresponding supervision signal. We use the same multi-scale training loss proposed in FlowNet [15]:

$$\mathcal{L}(\Theta) = \sum_{l=l_0}^{L} \alpha_l \sum_{\mathbf{x}} |\mathbf{w}_\Theta^l(\mathbf{x}) - \mathbf{w}_{GT}^l(\mathbf{x})|_2 + \gamma |\Theta|_2, \quad (3)$$

where $|\cdot|_2$ computes the L2 norm of a vector and the second term regularizes parameters of the model. For fine-tuning, we use the following robust training loss:

$$\mathcal{L}(\Theta) = \sum_{l=l_0}^{L} \alpha_l \sum_{\mathbf{x}} \left( |\mathbf{w}_\Theta^l(\mathbf{x}) - \mathbf{w}_{GT}^l(\mathbf{x})| + \epsilon \right)^q + \gamma |\Theta|_2 \quad (4)$$

where $|\cdot|$ denotes the L1 norm, $q<1$ gives less penalty to outliers, and $\epsilon$ is a small constant.

## 4. Experimental Results

**Implementation details.** The weights in the training loss (3) are set to be $\alpha_6 = 0.32, \alpha_5 = 0.08, \alpha_4 = 0.02, \alpha_3 = 0.01$, and $\alpha_2 = 0.005$. The trade-off weight $\gamma$ is set to be 0.0004. We scale the ground truth flow by 20 and downsample it to obtain the supervision signals at different levels. Note that we do not further scale the supervision signal at each level, the same as [15]. As a result, we need to scale the upsampled flow at each pyramid level for the warping layer. For example, at the second level, we scale the upsampled flow from the third level by a factor of 5 ($= 20/4$) before warping features of the second image. We use a 7-level pyramid and set $l_0$ to be 2, *i.e.*, our model outputs a quarter resolution optical flow and uses bilinear interpolation to obtain the full-resolution optical flow. We use a search range of 4 pixels to compute the cost volume at each level.

We first train the models using the FlyingChairs dataset in Caffe [28] using the $S_{long}$ learning rate schedule introduced in [24], *i.e.*, starting from 0.0001 and reducing the learning rate by half at 0.4M, 0.6M, 0.8M, and 1M iterations. The data augmentation scheme is the same as that in [24]. We crop $448 \times 384$ patches during data augmentation and use a batch size of 8. We then fine-tune the models on the FlyingThings3D dataset using the $S_{fine}$ schedule [24] while excluding image pairs with extreme motion (magnitude larger than 1000 pixels). The cropped image size is $768 \times 384$ and the batch size is 4. Finally, we fine-tune the models using the Sintel and KITTI training set and will explain the details below.

### 4.1. Main Results

**MPI Sintel.** When fine-tuning on Sintel, we crop $768 \times 384$ image patches, add horizontal flip, and remove additive noise during data augmentation. The batch size is 4. We use the robust loss function in Eq. (4) with $\epsilon = 0.01$ and $q = 0.4$. We test two schemes of fine-tuning. The first one, PWC-Net-ft, uses the clean and final passes of the Sintel training data throughout the fine-tuning process. The second one, PWC-Net-ft-final, uses only the final pass for the second half of fine-tuning. We test the second scheme because the DCFlow method learns the features using only the final pass of the training data. Thus we test the performance of PWC-Net when the final pass of the training data is given more weight.

At the time of writing, PWC-Net has lower average end-point error (EPE) than all published methods on the final pass of the MPI-Sintel benchmark (Table 1). It is the first time that an end-to-end method outperforms well-engineered and highly fine-tuned traditional methods on this benchmark. Further, PWC-Net is the fastest among all the top-performing methods (Fig. 1). We can further reduce the running time by dropping the DenseNet connections. The resulting PWC-Net-small model is about 5% less accurate but 40% faster than PWC-Net.

PWC-Net is less accurate than traditional approaches on the clean pass. Many traditional methods use image edges to refine motion boundaries, because the two are perfectly aligned in the clean pass. However, image edges in the final pass are corrupted by motion blur, atmospheric changes, and noise. Thus, the final pass is more realistic and challenging. The results on the final and clean sets suggest that PWC-Net may be better suited for real images, where the image edges are often corrupted.

PWC-Net has higher errors on the training set but lower errors on the test set than FlowNet2, suggesting that PWC-Net may have a more appropriate capacity for this task. Table 2 summarizes errors in different regions. PWC-Net performs relatively better in regions with large motion and away from the motion boundaries, probably because it has been trained using only data with large motion. Figure 4 shows the visual results of different variants of PWC-Net on the training and test sets of MPI Sintel. PWC-Net can recover sharp motion boundaries but may fail on small and rapidly moving objects, such as the left arm in "Market_5".

**KITTI.** When fine-tuning on KITTI, we crop $896 \times 320$ image patches and reduce the amount of rotation, zoom, and squeeze during data augmentation. The batch size is 4 too. The large patches can capture the large motion in the KITTI dataset. Since the ground truth is semi-dense, we upsample the predicted flow at the quarter resolution to compare with the scaled ground truth at the full resolution. We exclude the invalid pixels in computing the loss function.

At the time of writing, PWC-Net outperforms all published two-frame optical flow methods on the 2015 set, as shown in Table 3. It has the lowest percentage of flow outliers (Fl-all) in both all and non-occluded pixels (Table 4). PWC-Net has the second lowest percentage of outliers in non-occluded regions (Fl-noc) on the 2012 set, only inferior to SDF that assumes a rigidity constraint for the background. Although the rigidity assumption works well on the static scenes in the 2012 set, PWC-Net outperforms SDF in the 2015 set which mainly consists of dynamic scenes and is more challenging. The visual results in Fig. 5 qualitatively demonstrate the benefits of using the context network, DenseNet connections, and fine-tuning respectively. In particular, fine-tuning fixes large regions of errors in the test set, demonstrating the benefit of learning when the training and test data share similar statistics.

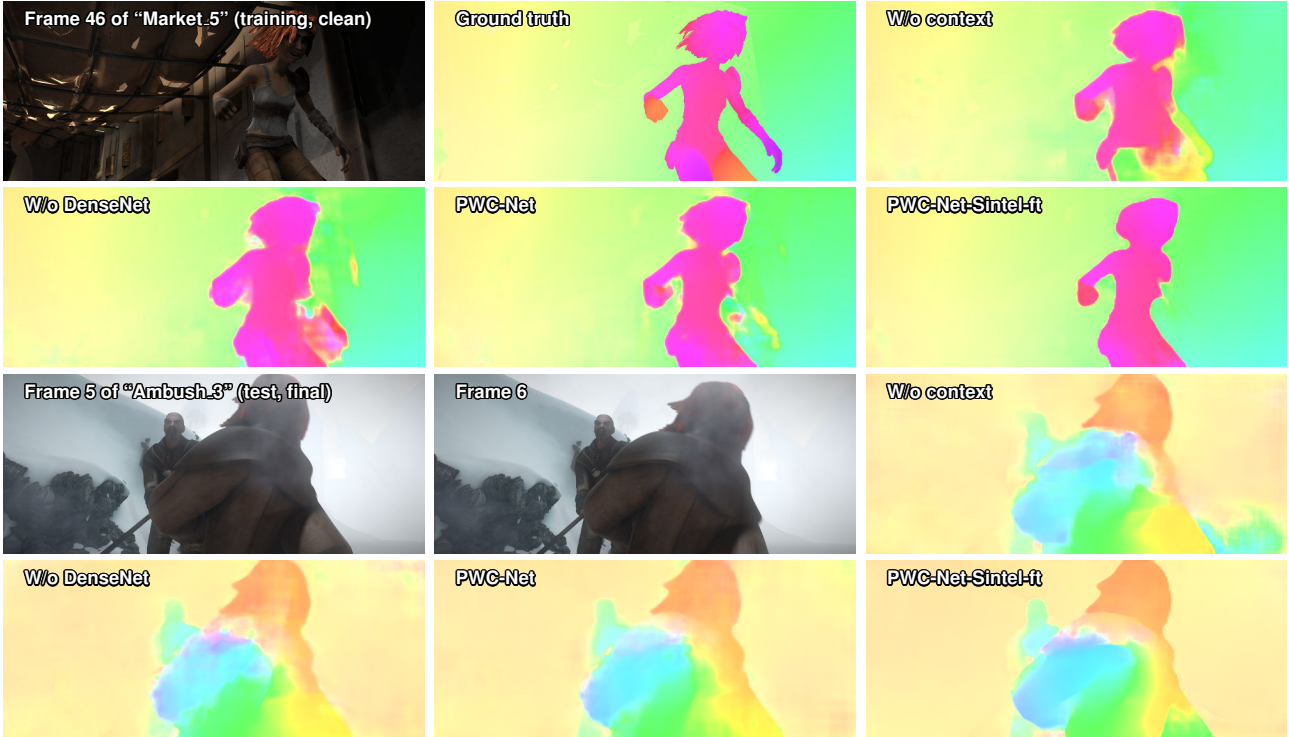

Figure 4. Results on Sintel *training* and *test* sets. Context network, DenseNet connections, and fine-tuning all improve the results.

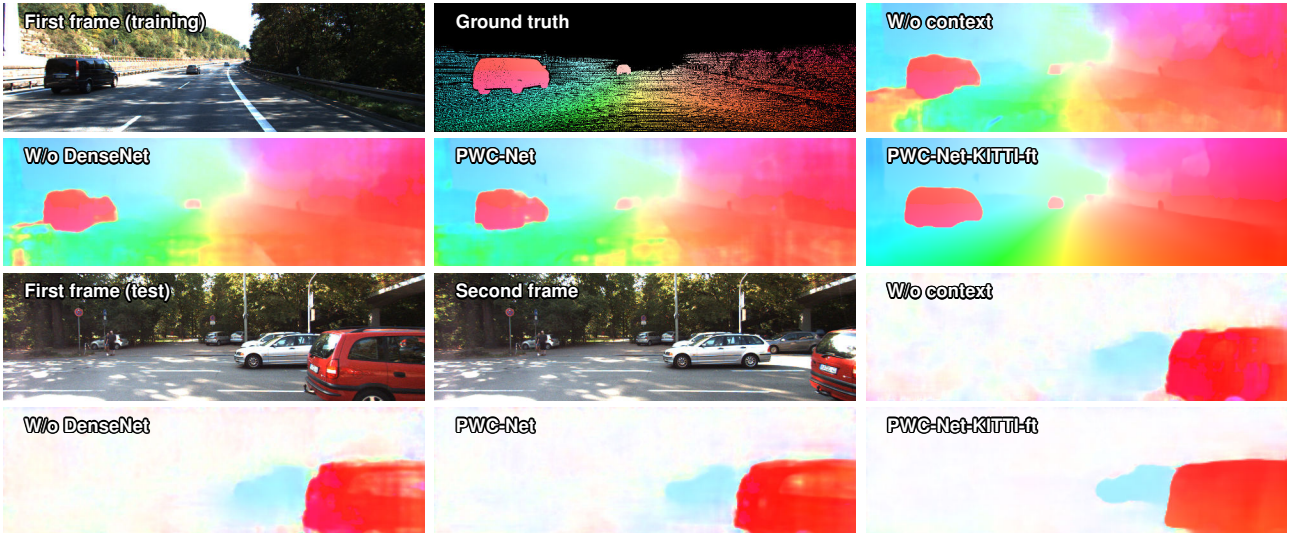

Figure 5. Results on KITTI 2015 *training* and *test* sets. Fine-tuning fixes large regions of errors and recovers sharp motion boundaries.

As shown in Table 4, FlowNet2 and PWC-Net have the most accurate results in the foreground regions, both outperforming the best published scene flow method, ISF [6].

Scene flow methods, however, have much lower errors in the static background region. The results suggest that synergizing advances in optical flow and scene flow could lead to more accurate results.

### 4.2. Ablation Experiments

**Feature pyramid extractor.** PWC-Net uses a two-layer CNN to extract features at each pyramid level. Table 5a summarizes the results of two variants that use one layer (↓) and three layers (↑) respectively. A larger-capacity feature get stuck at poor local minima, which can be detected by checking the validation errors after a few thousand iterations and fixed by running from a different random initialization.

Table 1. Average EPE results on MPI Sintel set. "-ft" means fine-tuning on the MPI Sintel *training* set and the numbers in the parenthesis are results on the data the methods have been fine-tuned on. ft-final gives more weight to the final pass during fine-tuning.

| Methods | Training | | Test | | Time |
| --- | --- | --- | --- | --- | --- |
| | Clean | Final | Clean | Final | (s) |
| PatchBatch [17] | - | - | 5.79 | 6.78 | 50.0 |
| EpicFlow [39] | - | - | 4.12 | 6.29 | 15.0 |
| CPM-flow [21] | - | - | 3.56 | 5.96 | 4.30 |
| FullFlow [14] | - | 3.60 | 2.71 | 5.90 | 240 |
| FlowFields [2] | - | - | 3.75 | 5.81 | 28.0 |
| MRFlow [53] | 1.83 | 3.59 | **2.53** | 5.38 | 480 |
| FlowFieldsCNN [3] | - | - | 3.78 | 5.36 | 23.0 |
| DCFlow [55] | - | - | 3.54 | 5.12 | 8.60 |
| SpyNet-ft [38] | (3.17) | (4.32) | 6.64 | 8.36 | 0.16 |
| FlowNet2.0 [24] | 2.02 | 3.14 | 3.96 | 6.02 | 0.12 |
| FlowNet2.0-ft [24] | (**1.45**) | (**2.01**) | 4.16 | 5.74 | 0.12 |
| PWC-Net-small | 2.83 | 4.08 | - | - | **0.02** |
| PWC-Net-small-ft | (2.27) | (2.45) | 5.05 | 5.32 | **0.02** |
| PWC-Net | 2.55 | 3.93 | - | - | 0.03 |
| PWC-Net-ft | (1.70) | (2.21) | 3.86 | 5.13 | 0.03 |
| PWC-Net-ft-final | (2.02) | (2.08) | 4.39 | **5.04** | 0.03 |

Table 2. Detailed results on the Sintel benchmark for different regions, velocities ($s$), and distances from motion boundaries ($d$).

| Final | matched | unmatched | $d_{0-10}$ | $d_{10-60}$ | $d_{60-140}$ | $s_{0-10}$ | $s_{10-40}$ | $s_{40+}$ |
| --- | --- | --- | --- | --- | --- | --- | --- | --- |
| PWC-Net | **2.44** | **27.08** | **4.68** | **2.08** | **1.52** | **0.90** | **2.99** | **31.28** |
| FlowNet2 | 2.75 | 30.11 | 4.82 | 2.56 | 1.74 | 0.96 | 3.23 | 35.54 |
| SpyNet | 4.51 | 39.69 | 6.69 | 4.37 | 3.29 | 1.40 | 5.53 | 49.71 |
| Clean | | | | | | | | |
| PWC-Net | **1.45** | **23.47** | 3.83 | **1.31** | **0.56** | 0.70 | 2.19 | **23.56** |
| FlowNet2 | 1.56 | 25.40 | **3.27** | 1.46 | 0.86 | **0.60** | **1.89** | 27.35 |
| SpyNet | 3.01 | 36.19 | 5.50 | 3.12 | 1.72 | 0.83 | 3.34 | 43.44 |

Table 3. Results on the KITTI dataset. "-ft" means fine-tuning on the KITTI *training* set and the numbers in the parenthesis are results on the data the methods have been fine-tuned on.

| Methods | KITTI 2012 | | | KITTI 2015 | | |
| --- | --- | --- | --- | --- | --- | --- |
| | AEPE | AEPE | Fl-Noc | AEPE | Fl-all | Fl-all |
| | train | test | test | train | train | test |
| EpicFlow [39] | - | 3.8 | 7.88% | - | - | 26.29 % |
| FullFlow [14] | - | - | - | - | - | 23.37 % |
| CPM-flow [21] | - | 3.2 | 5.79% | - | - | 22.40 % |
| PatchBatch [17] | - | 3.3 | 5.29% | - | - | 21.07% |
| FlowFields [2] | - | - | - | - | - | 19.80% |
| MRFlow [53] | - | - | - | - | 14.09 % | 12.19 % |
| DCFlow [55] | - | - | - | - | 15.09 % | 14.83 % |
| SDF [1] | - | 2.3 | **3.80**% | - | - | 11.01 % |
| MirrorFlow [23] | - | 2.6 | 4.38% | - | 9.93 % | 10.29% |
| SpyNet-ft [38] | (4.13) | 4.7 | 12.31% | - | - | 35.07% |
| FlowNet2 [24] | 4.09 | - | - | 10.06 | 30.37% | - |
| FlowNet2-ft [24] | (**1.28**) | 1.8 | 4.82% | (2.30) | (**8.61**%) | 10.41 % |
| PWC-Net | 4.14 | - | - | 10.35 | 33.67% | - |
| PWC-Net-ft | (1.45) | **1.7** | 4.22% | (**2.16**) | (9.80%) | **9.60**% |

Table 4. Detailed Results on the KITTI 2015 benchmark for the top three optical flow and two scene flow methods (below).

| Methods | Non-occluded pixels | | | All pixels | | |
| --- | --- | --- | --- | --- | --- | --- |
| | Fl-bg | Fl-fg | Fl-all | Fl-bg | Fl-fg | Fl-all |
| MirrorFlow [23] | 6.24% | 12.95% | 7.46% | **8.93**% | 17.07% | 10.29% |
| FlowNet2 [24] | 7.24% | **5.60**% | 6.94% | 10.75% | **8.75**% | 10.41% |
| PWC-Net | **6.14**% | 5.98% | **6.12**% | 9.66% | 9.31% | **9.60**% |
| OSF [37] | **4.21**% | 15.49% | 6.26% | 5.62% | 18.92% | 7.83% |
| ISF [6] | **4.21**% | 6.83% | **4.69**% | **5.40**% | 10.29% | **6.22**% |

Removing the context network results in larger errors on both the training and validation sets (Table 5c). Removing the DenseNet connections results in higher training error but lower validation errors when the model is trained on FlyingChairs. However, after the model is fine-tuned on FlyingThings3D, DenseNet leads to lower errors.

We also test a residual version of the optical flow estimator, which estimates a flow increment and adds it to the initial flow to obtain the refined flow. As shown in Table 5f, this residual version slightly improves the performance.

**Cost volume.** We test the search range to compute the cost volume, shown in Table 5d. A larger range leads to lower training error. However, all three settings have similar performance on Sintel, because a range of 2 at every level can already deal with a motion up to 200 pixels at the input resolution. A larger range has lower EPE on KITTI, likely because the images from the KITTI dataset have larger displacements than those from Sintel. A smaller range, however, seems to force the network to ignore pixels with extremely large motion and focus more on small-motion pixels, thereby achieving lower Fl-all scores.

**Warping.** Warping allows for estimating a small optical flow (increment) at each pyramid level to deal with a large optical flow. Removing the warping layers results in a significant loss of accuracy (Table 5e). Without the warping layer, PWC-Net still produces reasonable results, because the default search range of 4 to compute the cost volume is large enough to capture the motion of most sequences at the low-resolution pyramid levels.

**Dataset scheduling.** We also train PWC-Net using different dataset scheduling schemes, as shown in Table 6. Sequentially training on FlyingChairs, FlyingThings3D, and Sintel gradually improves the performance, consistent with the observations in [24]. Directly training using the test data leads to good "over-fitting" results, but the trained model does not perform as well on other datasets.

**Model size and running time.** Table 7 summarizes the model size for different CNN models. PWC-Net has about pyramid extractor leads to consistently better results on both the training and validation datasets.

**Optical flow estimator.** PWC-Net uses a five-layer CNN in the optical flow estimator at each level. Table 5b shows the results by two variants that use four layer ($\downarrow$) and seven layers ($\uparrow$) respectively. A larger-capacity optical flow estimator leads to better performance. However, we observe in our experiments that a deeper optical flow estimator might

|  | Chairs | Sintel Clean | Sintel Final | KITTI 2012 AEPE | KITTI 2012 Fl-all | KITTI 2015 AEPE | KITTI 2015 Fl-all |
|---|---|---|---|---|---|---|---|
| Full model | 2.00 | 3.33 | 4.59 | 5.14 | 28.67% | 13.20 | 41.79% |
| Feature ↑ | **1.92** | **3.03** | **4.17** | **4.57** | **26.73%** | **11.64** | **39.80%** |
| Feature ↓ | 2.18 | 3.36 | 4.56 | 5.75 | 30.79% | 14.05 | 44.92% |

(a) Larger-capacity **feature pyramid extractor** has better performance.

|  | Chairs | Sintel Clean | Sintel Final | KITTI 2012 AEPE | KITTI 2012 Fl-all | KITTI 2015 AEPE | KITTI 2015 Fl-all |
|---|---|---|---|---|---|---|---|
| Full model | 2.00 | 3.33 | 4.59 | 5.14 | 28.67% | 13.20 | 41.79% |
| Estimator ↑ | **1.92** | **3.09** | **4.50** | **4.64** | **25.34%** | **12.25** | **39.18%** |
| Estimator ↓ | 2.01 | 3.37 | 4.58 | 4.82 | 26.35% | 12.83 | 40.53% |

(b) Larger-capacity **optical flow estimator** has better performance.

|  | Trained on FlyingChairs | | | Fine-tuned on FlyingThings | | |
|---|---|---|---|---|---|---|
|  | Chairs | Clean | Final | Chairs | Clean | Final |
| Full model | **2.00** | 3.33 | 4.59 | **2.34** | **2.60** | **3.95** |
| No DenseNet | 2.06 | **3.09** | **4.37** | 2.48 | 2.83 | 4.08 |
| No Context | 2.23 | 3.47 | 4.74 | 2.55 | 2.75 | 4.13 |

(c) **Context network** consistently helps; **DenseNet** helps after fine-tuning.

| Max. Disp. | Chairs | Sintel Clean | Sintel Final | KITTI 2012 AEPE | KITTI 2012 Fl-all | KITTI 2015 AEPE | KITTI 2015 Fl-all |
|---|---|---|---|---|---|---|---|
| Full model | 2.00 | 3.33 | 4.59 | 5.14 | 28.67% | 13.20 | 41.79% |
| 2 | 2.09 | **3.30** | **4.50** | 5.26 | **25.99%** | 13.67 | **38.99%** |
| 6 | **1.97** | 3.31 | 4.60 | **4.96** | 27.05% | **12.97** | 40.94% |

(d) **Cost volume.** PWC-Net can handle large motion with small search range.

|  | Chairs | Sintel Clean | Sintel Final | KITTI 2012 AEPE | KITTI 2012 Fl-all | KITTI 2015 AEPE | KITTI 2015 Fl-all |
|---|---|---|---|---|---|---|---|
| Full model | **2.00** | **3.33** | **4.59** | **5.14** | **28.67%** | **13.20** | **41.79%** |
| No warping | 2.17 | 3.79 | 5.30 | 5.80 | 32.73% | 13.74 | 44.87% |

(e) **Warping layer** is a critical component for the performance.

|  | Chairs | Sintel Clean | Sintel Final | KITTI 2012 AEPE | KITTI 2012 Fl-all | KITTI 2015 AEPE | KITTI 2015 Fl-all |
|---|---|---|---|---|---|---|---|
| Full model | 2.00 | 3.33 | 4.59 | 5.14 | 28.67% | 13.20 | 41.79% |
| Residual | **1.96** | **3.14** | **4.43** | **4.87** | **27.74%** | **12.58** | **41.16%** |

(f) **Residual connections** in the optical flow estimator are helpful.

Table 5. **Ablation experiments**. Unless explicitly stated, the models have been trained on the FlyingChairs dataset.

Table 6. **Training dataset schedule** leads to better local minima. () indicates results on the dataset the method has been trained on.

| Data | Chairs AEPE | Sintel (AEPE) Clean | Final | KITTI 2012 AEPE | KITTI 2012 Fl-all | KITTI 2015 AEPE | KITTI 2015 Fl-all |
|---|---|---|---|---|---|---|---|
| Chairs | (**2.00**) | 3.33 | 4.59 | 5.14 | 28.67% | 13.20 | 41.79% |
| Chairs-Things | 2.30 | 2.55 | 3.93 | 4.14 | 21.38% | 10.35 | 33.67% |
| Chairs-Things-Sintel | 2.56 | (**1.70**) | (**2.21**) | **2.94** | **12.70%** | **8.15** | **24.35%** |
| Sintel | 3.69 | (1.86) | (2.31) | 3.68 | 16.65% | 10.52 | 30.49% |

17 times fewer parameters than FlowNet2. PWC-Net-small further reduces this by an additional 2 times via dropping DenseNet connections and is more suitable for memory-limited applications.

The timings have been obtained on the same desktop with an NVIDIA Pascal TitanX GPU. For more precise timing, we exclude the reading and writing time when benchmarking the forward and backward inference time. PWC-Net is about 2 times faster in forward inference and at least 3 times faster in training than FlowNet2.

Table 7. **Model size and running time.** PWC-Net-small drops DenseNet connections. For training, the lower bound of 14 days for FlowNet2 is obtained by 6(FlowNetC) + 2×4 (FlowNetS).

| Methods | FlowNetS | FlowNetC | FlowNet2 | SpyNet | PWC-Net | PWC-Net-small |
|---|---|---|---|---|---|---|
| #parameters (M) | 38.67 | 39.17 | 162.49 | 1.2 | 8.75 | 4.08 |
| Parameter Ratio | 23.80% | 24.11% | 100% | 0.74% | 5.38% | 2.51% |
| Memory (MB) | 154.5 | 156.4 | 638.5 | 9.7 | 41.1 | 22.9 |
| Memory Ratio | 24.20% | 24.49% | 100% | 1.52% | 6.44% | 3.59% |
| Training (days) | 4 | 6 | >14 | - | 4.8 | 4.1 |
| Forward (ms) | 11.40 | 21.69 | 84.80 | - | 28.56 | 20.76 |
| Backward (ms) | 16.71 | 48.67 | 78.96 | - | 44.37 | 28.44 |

**Discussions.** Both PWC-Net and SpyNet have been inspired by classical principles for flow and stereo but have significant differences. SpyNet uses image pyramids while PWC-Net learns feature pyramids. SpyNet feeds CNNs with images, while PWC-Net feeds a cost volume. As the cost volume is a more discriminative representation of the search space for optical flow, the learning task for CNNs becomes easier. Regarding performance, PWC-Net outperforms SpyNet by a significant margin. Additionally, SpyNet has been trained sequentially, while PWC-Net can be trained end-to-end from scratch.

FlowNet2 [24] achieves impressive performance by stacking several basic models into a large-capacity model. The much smaller PWC-Net obtains similar or better performance by embedding classical principles into the network architecture. It would be interesting to use PWC-Net as a building block to design large networks.

## 5. Conclusions

We have developed a compact but effective CNN model for optical flow using simple and well-established principles: pyramidal processing, warping, and the use of a cost volume. Combining deep learning with domain knowledge not only reduces the model size but also improves the performance. PWC-Net is about 17 times smaller in size, 2 times faster in inference, and easier to train than FlowNet2. It outperforms all published optical flow methods to date on the Sintel final pass and KITTI 2015 benchmarks, running at about 35 fps on Sintel resolution ($1024 \times 436$) images.

Given the compactness, efficiency, and effectiveness of PWC-Net, we expect it to be a useful component of many video processing systems. To enable comparison and further innovations, we make our models available on our project website.

**Acknowledgements** We would like to thank Eddy Ilg for clarifying details about the FlowNet2 paper, Ming-Hsuan Yang for helpful suggestions, Michael Pellauer for proofreading, and the anonymous reviewers for constructive comments.

In this appendix, Section 1 provides more ablation and visual results. Section 2 summarizes the details of our network. Section 3 shows the screenshot of the MPI Sintel final pass, KITTI 2012, and KITTI 2015 public tables at the time of submission (November 15th, 2017). Section 4 shows the learned features at the first level of the feature pyramid extractor.

## 1. More Ablation and Visual Results

Figure 1 shows the enlarged images of Figure 1 in the main manuscript. PWC-Net outperforms all published methods on the MPI Sintel final pass benchmark in both accuracy and running time. It also reaches the best balance between size and accuracy among existing end-to-end CNN models.

Table 1 shows more ablation results, in particular, the full results for models trained on FlyingChairs (Table 1a) and then fine-tuned on FlyingThings3D (Table 1b). To further test the dilated convolutions, we replace the dilated convolutions of the context network with plain convolutions. Using plain convolutions has worse performance on Chairs and Sintel, and is slightly better on KITTI. We also have independent runs of the same PWC-Net that only differ in the random initialization. As shown in Table 1d, the two independent runs lead to models that have close performances, although not exactly the same.

Figures 2 and 3 provide more visual results by PWC-Net on the MPI Sintel final pass and KITTI 2015 test sets. PWC-Net can recover sharp motion boundaries in the presence of large motion, severe occlusions, and strong shadow and atmospheric effects. However, PWC-Net tends to produce errors on objects with thin structures that rarely occur in the training set, such as the wheels of the bicycle in the third row of Figure 3.

## 2. Network Details

Figure 4 shows the architecture for the 7-level feature pyramid extractor network used in our experiment. Note that the bottom level consists of the original input images. Figure 5 shows the optical flow estimator network at pyramid level 2. The optical flow estimator networks at other levels have the same structure except for the top level, which does not have the upsampled optical flow and directly computes cost volume using features of the first and second images. Figure 6 shows the context network that is adopted only at pyramid level 2.

## 3. Screenshots of MPI Sintel and KITTI Public Table

Figures 7-9 respectively show the screenshots of the MPI Sintel final pass, KITTI 2015, and KITTI 2012 public tables at the time of submission (November 15th, 2017). Among all optical flow methods, PWC-Net is ranked 1st on both MPI Sintel final and KITTI 2015, and 2nd on KITTI 2012. Note that the 1st-ranked method on KITTI 2012, SDF [1], assumes a rigidity constraint for the background, which is well-suited to the static scenes in KITTI 2012. PWC-Net performs better than SDF on KITTI 2015 that contains dynamic objects and is more challenging.

## 4. Learned Features

Figure 10 shows the learned filters for the first convolution layer by PWC-Net and the feature responses to an input image. These filters tend to focus on regions of different properties in the input image. After training on FlyingChairs, fine-tuning on FlyingThings3D and Sintel does not change these filters much.

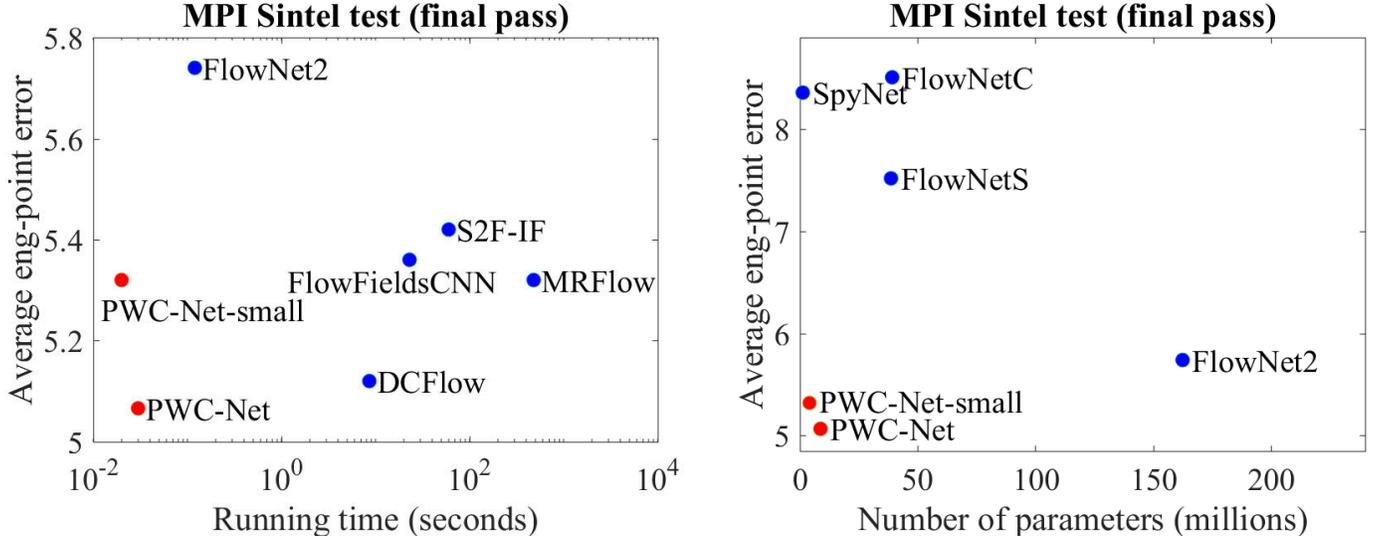

Figure 1. Left: PWC-Net outperforms all published methods on the MPI Sintel final pass benchmark in both accuracy and running time. Right: PWC-Net reaches the best balance between size and accuracy among existing end-to-end CNN models.

|  | Chairs | Sintel Clean | Sintel Final | KITTI 2012 AEPE | KITTI 2012 Fl-all | KITTI 2015 AEPE | KITTI 2015 Fl-all |
|---|---|---|---|---|---|---|---|
| Full model | **2.00** | 3.33 | 4.59 | 5.14 | 28.67% | 13.20 | 41.79% |
| No context | 2.06 | **3.09** | **4.37** | **4.77** | **25.35%** | **12.03** | **39.21%** |
| No DenseNet | 2.23 | 3.47 | 4.74 | 5.63 | 28.53% | 14.02 | 40.33% |
| Neither | 2.22 | 3.15 | 4.49 | 5.46 | 28.02% | 13.14 | 40.03% |

(a) **Trained on FlyingChairs.**

|  | Chairs | Sintel Clean | Sintel Final | KITTI 2012 AEPE | KITTI 2012 Fl-all | KITTI 2015 AEPE | KITTI 2015 Fl-all |
|---|---|---|---|---|---|---|---|
| Full model | **2.30** | **2.55** | **3.93** | **4.14** | **21.38%** | **10.35** | **33.67%** |
| No context | 2.48 | 2.82 | 4.09 | 4.39 | 21.91% | 10.82 | 34.44% |
| No DenseNet | 2.54 | 2.72 | 4.09 | 4.91 | 24.04% | 11.52 | 34.79% |
| Neither | 2.65 | 2.83 | 4.24 | 4.89 | 24.52% | 12.01 | 35.73% |

(b) **Fine-tuend on FlyingThings3D after FlyingChairs.**

|  | Chairs | Sintel Clean | Sintel Final | KITTI 2012 AEPE | KITTI 2012 Fl-all | KITTI 2015 AEPE | KITTI 2015 Fl-all |
|---|---|---|---|---|---|---|---|
| Dilated conv | **2.00** | **3.33** | **4.59** | 5.14 | 28.67% | 13.20 | 41.79% |
| Plain conv | 2.03 | 3.39 | 4.85 | 5.29 | **25.86%** | 13.17 | **38.67%** |

(c) **Dilated vs plain convolutions** for the context network.

|  | Chairs | Sintel Clean | Sintel Final | KITTI 2012 AEPE | KITTI 2012 Fl-all | KITTI 2015 AEPE | KITTI 2015 Fl-all |
|---|---|---|---|---|---|---|---|
| Run 1 | 2.00 | 3.33 | **4.59** | 5.14 | 28.67% | 13.20 | 41.79% |
| Run 2 | **2.00** | **3.33** | 4.65 | **4.81** | **27.12%** | **13.10** | **40.84%** |

(d) **Two independent runs** result in slightly different models.

Table 1. **More ablation experiments**. Unless explicitly stated, the models have been trained on the FlyingChairs dataset.

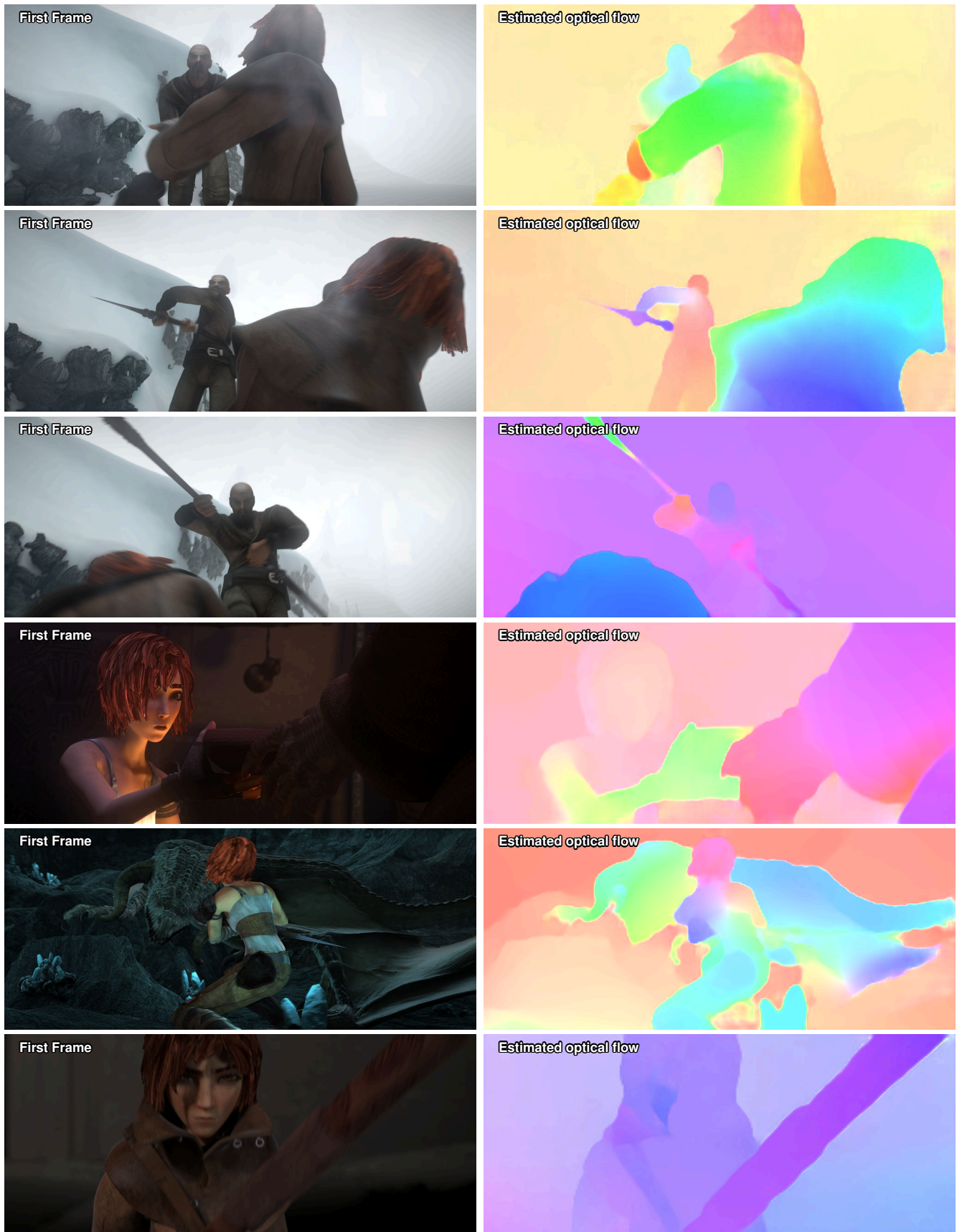

Figure 2. More PWC-Net results on the MPI Sintel final pass dataset.

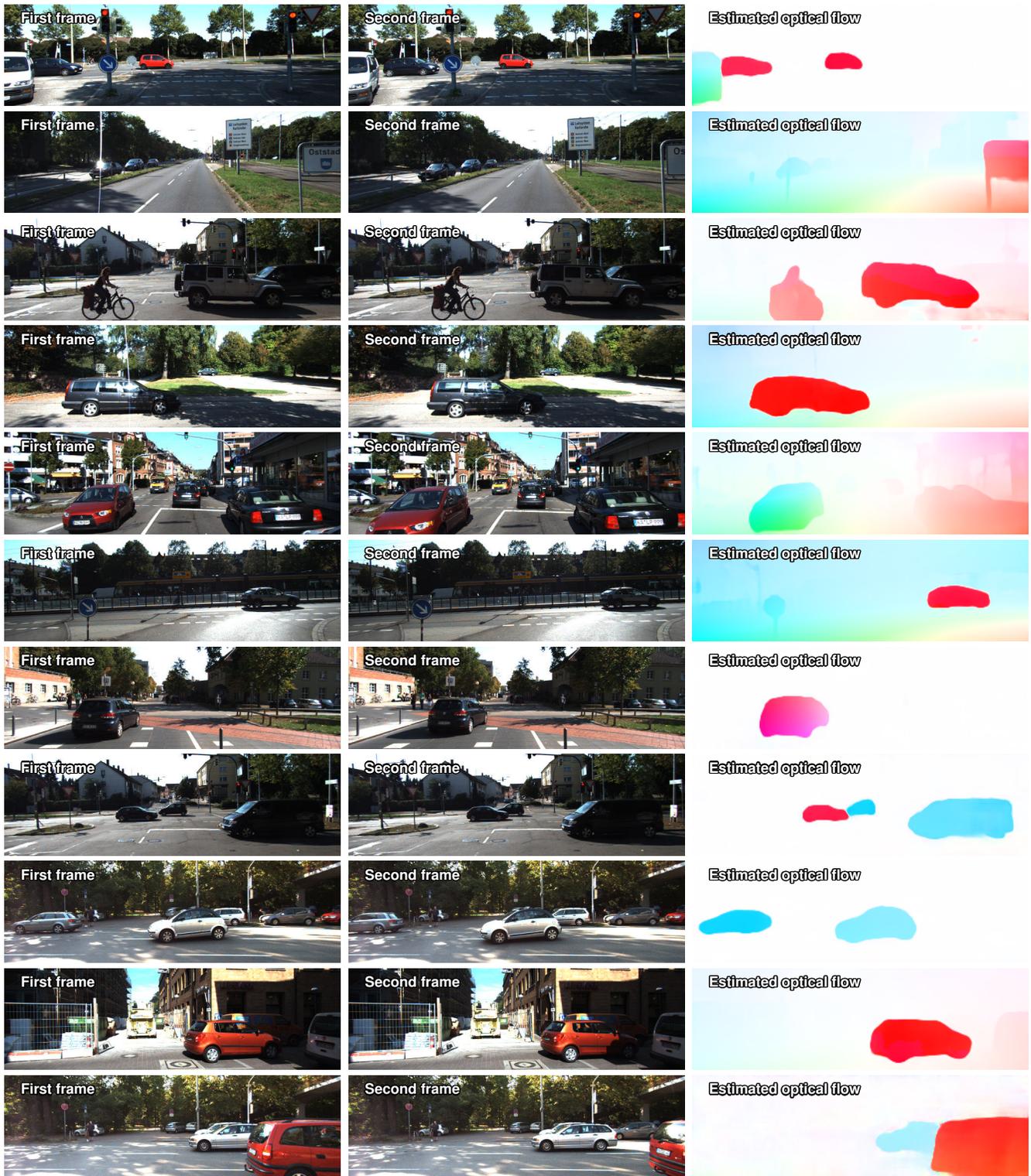

Figure 3. More PWC-Net results on KITTI 2015 test set. PWC-Net can recover sharp motion boundaries despite large motion, strong shadows, and severe occlusions. Thin structures, such as the bicycle, are challenging to PWC-Net, probably because the training set has no training samples of bicycles.

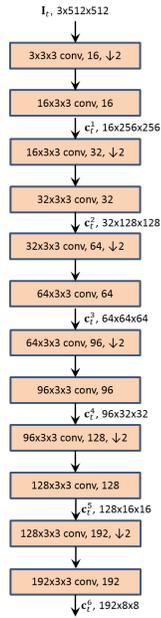

Figure 4. The feature pyramid extractor network. The first image ($t = 1$) and the second image ($t = 2$) are encoded using the same Siamese network. Each convolution is followed by a leaky ReLU unit. The convolutional layer and the $\times 2$ downsampling layer at each level is implemented using a single convolutional layer with a stride of 2. $\mathbf{c}_t^l$ denotes extracted features of image $t$ at level $l$;

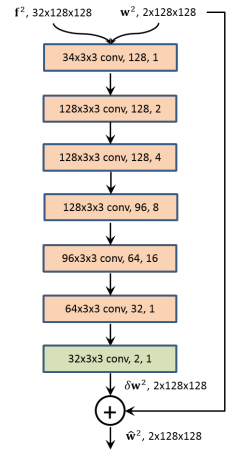

Figure 6. The context network at pyramid level 2. Each convolutional layer is followed by a leaky ReLU unit except the last (light green) one that outputs the optical flow. The last number in each convolutional layer denotes the dilation constant.

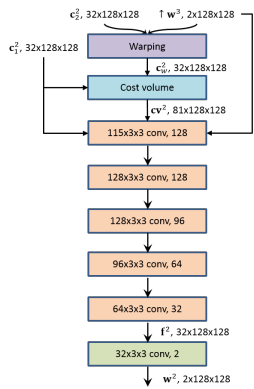

Figure 5. The optical flow estimator network at pyramid level 2. Each convolutional layer is followed by a leaky ReLU unit except the last (light green) one that outputs the optical flow.

|  | EPE all | EPE matched | EPE unmatched | d0-10 | d10-60 | d60-140 | s0-10 | s10-40 | s40+ |  |
|---|---|---|---|---|---|---|---|---|---|---|
| GroundTruth [1] | 0.000 | 0.000 | 0.000 | 0.000 | 0.000 | 0.000 | 0.000 | 0.000 | 0.000 | Visualize Results |
| PWC-Net [2] | 5.042 | 2.445 | 26.221 | 4.636 | 2.087 | 1.475 | 0.799 | 2.986 | 31.070 | Visualize Results |
| DCFlow [3] | 5.119 | 2.283 | 28.228 | 4.665 | 2.108 | 1.440 | 1.052 | 3.434 | 29.351 | Visualize Results |
| FlowFieldsCNN [4] | 5.363 | 2.303 | 30.313 | 4.718 | 2.020 | 1.399 | 1.032 | 3.065 | 32.422 | Visualize Results |
| MR-Flow [5] | 5.376 | 2.818 | 26.235 | 5.109 | 2.395 | 1.755 | 0.908 | 3.443 | 32.221 | Visualize Results |
| FTFlow [6] | 5.390 | 2.268 | 30.841 | 4.513 | 1.964 | 1.366 | 1.046 | 3.322 | 31.936 | Visualize Results |
| S2F-IF [7] | 5.417 | 2.549 | 28.795 | 4.745 | 2.198 | 1.712 | 1.157 | 3.468 | 31.262 | Visualize Results |
| InterpoNet_ff [8] | 5.535 | 2.372 | 31.296 | 4.720 | 2.018 | 1.532 | 1.064 | 3.496 | 32.633 | Visualize Results |
| PGM-C [9] | 5.591 | 2.672 | 29.389 | 4.975 | 2.340 | 1.791 | 1.057 | 3.421 | 33.339 | Visualize Results |
| RicFlow [10] | 5.620 | 2.765 | 28.907 | 5.146 | 2.366 | 1.679 | 1.088 | 3.364 | 33.573 | Visualize Results |
| InterpoNet_cpm [11] | 5.627 | 2.594 | 30.344 | 4.975 | 2.213 | 1.640 | 1.042 | 3.575 | 33.321 | Visualize Results |

Figure 7. Screenshot of the MPI Sintel final pass public table. PWC-Net has the lowest average end-point error (EPE) among all evaluated methods as of November 15th, 2017.

|  | Method | Setting | Code | Fl-bg | Fl-fg | Fl-all | Density | Runtime | Environment | Compare |
|---|---|---|---|---|---|---|---|---|---|---|
| 1 | PSPO |  |  | 4.35 % | 15.21 % | 6.15 % | 100.00 % | 5 min | 1 core @ 2.5 Ghz (Matlab + C/C++) |  |
| 2 | ISF |  |  | 5.40 % | 10.29 % | 6.22 % | 100.00 % | 10 min | 1 core @ 3 Ghz (C/C++) |  |

A. Behl, O. Jafari, S. Mustikovela, H. Alhaija, C. Rother and A. Geiger: Bounding Boxes, Segmentations and Object Coordinates: How Important is Recognition for 3D Scene Flow Estimation in Autonomous Driving Scenarios?. International Conference on Computer Vision (ICCV) 2017.

| 3 | PRSM |  | code | 5.33 % | 13.40 % | 6.68 % | 100.00 % | 300 s | 1 core @ 2.5 Ghz (C/C++) |  |

C. Vogel, K. Schindler and S. Roth: 3D Scene Flow Estimation with a Piecewise Rigid Scene Model. ijcv 2015.

| 4 | OSF+TC |  |  | 5.76 % | 13.31 % | 7.02 % | 100.00 % | 50 min | 1 core @ 2.5 Ghz (C/C++) |  |

M. Neoral and J. Šochman: Object Scene Flow with Temporal Consistency. 22nd Computer Vision Winter Workshop (CVWW) 2017.

| 5 | SSF |  |  | 5.63 % | 14.71 % | 7.14 % | 100.00 % | 5 min | 1 core @ 2.5 Ghz (Matlab + C/C++) |  |

Z. Ren, D. Sun, J. Kautz and E. Sudderth: Cascaded Scene Flow Prediction using Semantic Segmentation. International Conference on 3D Vision (3DV) 2017.

| 6 | SOSF |  |  | 5.42 % | 17.24 % | 7.39 % | 100.00 % | 55 min | 1 core @ 2.5 Ghz (Matlab + C/C++) |  |
| 7 | OSF |  | code | 5.62 % | 18.92 % | 7.83 % | 100.00 % | 50 min | 1 core @ 2.5 Ghz (C/C++) |  |

M. Menze and A. Geiger: Object Scene Flow for Autonomous Vehicles. Conference on Computer Vision and Pattern Recognition (CVPR) 2015.

| 8 | PWC-Net |  |  | 9.66 % | 9.31 % | 9.60 % | 100.00 % | 0.03 s | NVIDIA Pascal Titan X |  |
| 9 | MirrorFlow |  |  | 8.93 % | 17.07 % | 10.29 % | 100.00 % | 11 min | 4 core @ 2.2 Ghz (C/C++) |  |

J. Hur and S. Roth: MirrorFlow: Exploiting Symmetries in Joint Optical Flow and Occlusion Estimation. ICCV 2017.

| 10 | FlowNet2 |  |  | 10.75 % | 8.75 % | 10.41 % | 100.00 % | 0.12 s | GPU Nvidia GeForce GTX 1080 |  |
| 11 | SDF |  |  | 8.61 % | 23.01 % | 11.01 % | 100.00 % | TBA | 1 core @ 2.5 Ghz (C/C++) |  |

M. Bai*, W. Luo*, K. Kundu and R. Urtasun: Exploiting Semantic Information and Deep Matching for Optical Flow. ECCV 2016.

| 12 | UnFlow |  |  | 10.15 % | 15.93 % | 11.11 % | 100.00 % | 0.12 s | GPU @ 1.5 Ghz (Python + C/C++) |  |

S. Meister, J. Hur and S. Roth: UnFlow: Unsupervised Learning of Optical Flow with a Bidirectional Census Loss. AAAI 2018.

| 13 | FSF+MS |  |  | 8.48 % | 25.43 % | 11.30 % | 100.00 % | 2.7 s | 4 cores @ 3.5 Ghz (C/C++) |  |

T. Taniai, S. Sinha and Y. Sato: Fast Multi-frame Stereo Scene Flow with Motion Segmentation. IEEE Conference on Computer Vision and Pattern Recognition (CVPR 2017) 2017.

| 14 | CNNF+PMBP |  |  | 10.08 % | 18.56 % | 11.49 % | 100.00 % | 45 min | 1 cores @ 3.5 Ghz (C/C++) |  |
| 15 | MR-Flow |  | code | 10.13 % | 22.51 % | 12.19 % | 100.00 % | 8 min | 1 core @ 2.5 Ghz (Python + C/C++) |  |

J. Wulff, L. Sevilla-Lara and M. Black: Optical Flow in Mostly Rigid Scenes. IEEE Conf. on Computer Vision and Pattern Recognition (CVPR) 2017.

Figure 8. Screenshot of the KITTI 2015 public table. PWC-Net has the lowest percentage of error (Fl-all) among all optical flow methods, only inferior to scene flow methods that use additional stereo input information.

| | Method | Setting | Code | Out-Noc | Out-All | Avg-Noc | Avg-All | Density | Runtime | Environment | Compare |
|---|---|---|---|---|---|---|---|---|---|---|---|
| 1 | PRSM | | code | 2.46 % | 4.23 % | 0.7 px | 1.0 px | 100.00 % | 300 s | 1 core @ 2.5 Ghz (Matlab + C/C++) | |
| | C. Vogel, K. Schindler and S. Roth: 3D Scene Flow Estimation with a Piecewise Rigid Scene Model. ijcv 2015. | | | | | | | | | | |
| 2 | VC-SF | | | 2.72 % | 4.84 % | 0.8 px | 1.3 px | 100.00 % | 300 s | 1 core @ 2.5 Ghz (Matlab + C/C++) | |
| | C. Vogel, S. Roth and K. Schindler: View-Consistent 3D Scene Flow Estimation over Multiple Frames. Proceedings of European Conference on Computer Vision. Lecture Notes in, Computer Science 2014. | | | | | | | | | | |
| 3 | SPS-StFl | | | 2.82 % | 5.61 % | 0.8 px | 1.3 px | 100.00 % | 35 s | 1 core @ 3.5 Ghz (C/C++) | |
| | K. Yamaguchi, D. McAllester and R. Urtasun: Efficient Joint Segmentation, Occlusion Labeling, Stereo and Flow Estimation. ECCV 2014. | | | | | | | | | | |
| 4 | SPS-Fl | | | 3.38 % | 10.06 % | 0.9 px | 2.9 px | 100.00 % | 11 s | 1 core @ 3.5 Ghz (C/C++) | |
| | K. Yamaguchi, D. McAllester and R. Urtasun: Efficient Joint Segmentation, Occlusion Labeling, Stereo and Flow Estimation. ECCV 2014. | | | | | | | | | | |
| 5 | OSF | | code | 3.47 % | 6.34 % | 1.0 px | 1.5 px | 100.00 % | 50 min | 1 core @ 3.0 Ghz (Matlab + C/C++) | |
| | M. Menze and A. Geiger: Object Scene Flow for Autonomous Vehicles. Conference on Computer Vision and Pattern Recognition (CVPR) 2015. | | | | | | | | | | |
| 6 | PR-Sf+E | | | 3.57 % | 7.07 % | 0.9 px | 1.6 px | 100.00 % | 200 s | 4 cores @ 3.0 Ghz (Matlab + C/C++) | |
| | C. Vogel, K. Schindler and S. Roth: Piecewise Rigid Scene Flow. International Conference on Computer Vision (ICCV) 2013. | | | | | | | | | | |
| 7 | PCBP-Flow | | | 3.64 % | 8.28 % | 0.9 px | 2.2 px | 100.00 % | 3 min | 4 cores @ 2.5 Ghz (Matlab + C/C++) | |
| | K. Yamaguchi, D. McAllester and R. Urtasun: Robust Monocular Epipolar Flow Estimation. CVPR 2013. | | | | | | | | | | |
| 8 | PR-Sceneflow | | | 3.76 % | 7.39 % | 1.2 px | 2.8 px | 100.00 % | 150 sec | 4 core @ 3.0 Ghz (Matlab + C/C++) | |
| | C. Vogel, K. Schindler and S. Roth: Piecewise Rigid Scene Flow. International Conference on Computer Vision (ICCV) 2013. | | | | | | | | | | |
| 9 | SDF | | | 3.80 % | 7.69 % | 1.0 px | 2.3 px | 100.00 % | TBA s | 1 core @ 2.5 Ghz (C/C++) | |
| | M. Bai*, W. Luo*, K. Kundu and R. Urtasun: Exploiting Semantic Information and Deep Matching for Optical Flow. ECCV 2016. | | | | | | | | | | |
| 10 | MotionSLIC | | | 3.91 % | 10.56 % | 0.9 px | 2.7 px | 100.00 % | 11 s | 1 core @ 3.0 Ghz (C/C++) | |
| | K. Yamaguchi, D. McAllester and R. Urtasun: Robust Monocular Epipolar Flow Estimation. CVPR 2013. | | | | | | | | | | |
| 11 | PWC-Net | | | 4.22 % | 8.10 % | 0.9 px | 1.7 px | 100.00 % | 0.03 s | NVIDIA Pascal Titan X | |
| 12 | TBR | | | 4.24 % | 7.50 % | 0.9 px | 1.5 px | 100.00 % | 1750 s | 4 cores @ 2.5 Ghz (Matlab + C/C++) | |
| 13 | UnFlow | | | 4.28 % | 8.42 % | 0.9 px | 1.7 px | 100.00 % | 0.12 s | GPU @ 1.5 Ghz (Python + C/C++) | |
| | S. Meister, J. Hur and S. Roth: UnFlow: Unsupervised Learning of Optical Flow with a Bidirectional Census Loss. AAAI 2018. | | | | | | | | | | |
| 14 | MirrorFlow | | | 4.38 % | 8.20 % | 1.2 px | 2.6 px | 100.00 % | 11 min | 4 core @ 2.2 Ghz (C/C++) | |
| | J. Hur and S. Roth: MirrorFlow: Exploiting Symmetries in Joint Optical Flow and Occlusion Estimation. ICCV 2017. | | | | | | | | | | |

Figure 9. Screenshot of the KITTI 2012 public table. SDF [1] is the only optical flow method that has lower percentage of outliers in non-occluded regions (Out-Noc) than PWC-Net. However, SDF assumes a rigidity constraint for the background, which is well-suited for the static scenes in the KITTI 2012 set.

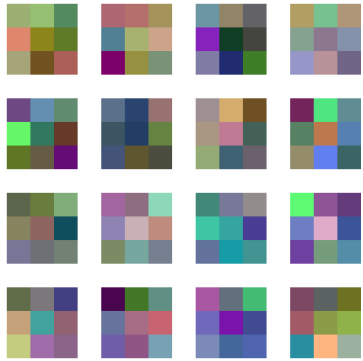

(a) Learned filters

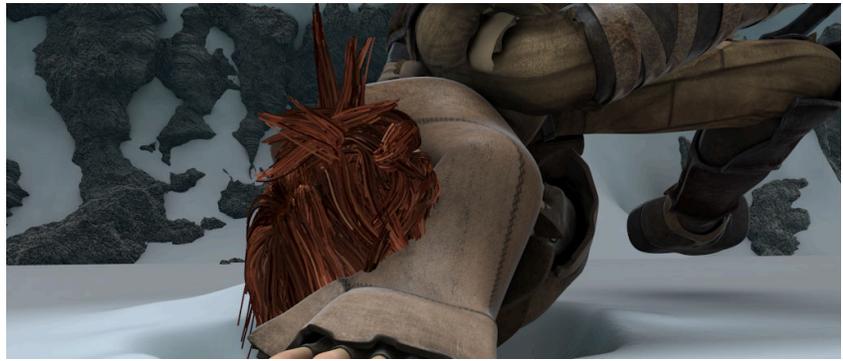

(b) Input

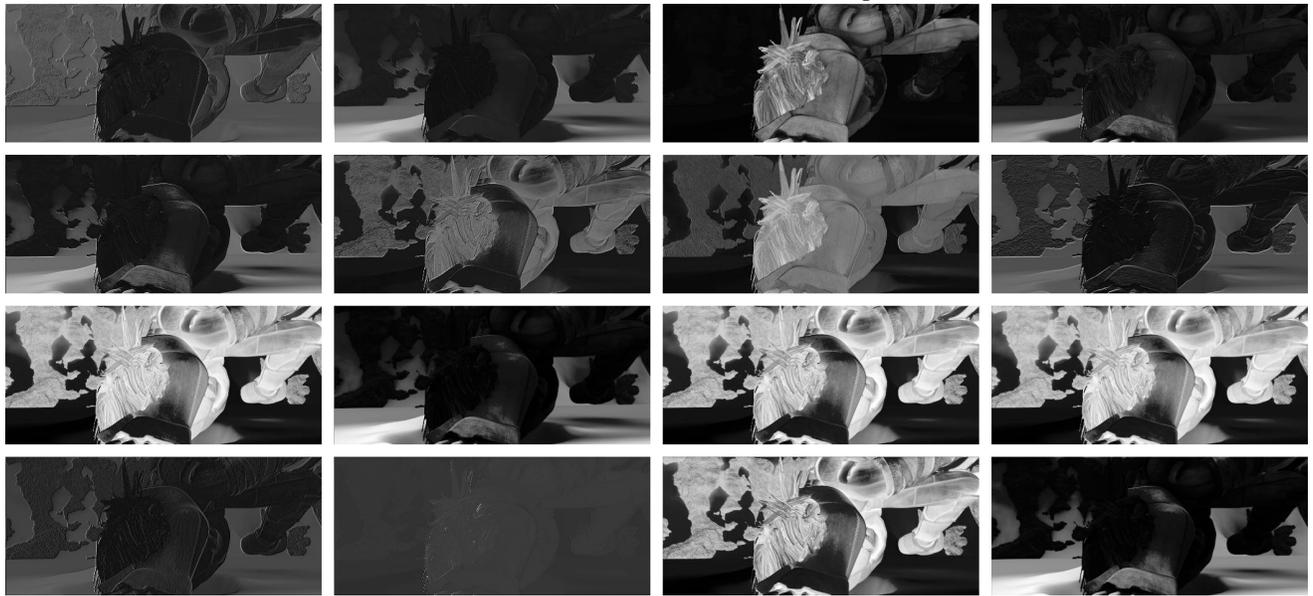

(c) Feature responses

Figure 10. Learned filters at the first convolutional layer of PWC-Net and the filter responses to an input image.